\documentclass{article}%
\usepackage[T1]{fontenc}%
\usepackage[utf8]{inputenc}%
\usepackage{lmodern}%
\usepackage{textcomp}%
\usepackage{lastpage}%
\usepackage{array}
\usepackage{url}
\usepackage{geometry}%
\usepackage{color,soul}
\usepackage{multicol}%
\usepackage{graphicx}%
\usepackage{lipsum}%
\usepackage{caption}%
\usepackage{booktabs}%
\usepackage{ragged2e}%
\usepackage{placeins}%
\usepackage{breqn}  
\usepackage{fancyvrb} 

\title{$AI^{2}$: The next leap toward native language based and explainable machine learning framework}%
\author{Jean{-}Sébastien Dessureault, Daniel Massicotte}%
\date{\today}%
\begin{document}%
\normalsize%
\maketitle%
\justify%

\section*{ABSTRACT}%
\label{sec:ABSTRACT}%
The machine learning frameworks flourished in the last decades, allowing artificial intelligence to get out of academic circles to be applied to enterprise domains. This field has significantly advanced, but there is still some meaningful improvement to reach the subsequent expectations. The proposed framework, named AI$^{2}$, uses a natural language interface that allows a non-specialist to benefit from machine learning algorithms without necessarily knowing how to program with a programming language. The primary contribution of the AI$^{2}$ framework allows a user to call the machine learning algorithms in English, making its interface usage easier. The second contribution is greenhouse gas (GHG) awareness. It has some strategies to evaluate the GHG generated by the algorithm to be called and to propose alternatives to find a solution without executing the energy-intensive algorithm. Another contribution is a preprocessing module that helps to describe and to load data properly. Using an English text-based chatbot, this module guides the user to define every dataset so that it can be described, normalized, loaded and divided appropriately. The last contribution of this paper is about explainability. For decades, the scientific community has known that machine learning algorithms imply the famous black-box problem. Traditional machine learning methods convert an input into an output without being able to justify this result. The proposed framework explains the algorithm's process with the proper texts, graphics and tables. The results, declined in five cases, present usage applications from the user's English command to the explained output. Ultimately, the AI$^{2}$ framework represents the next leap toward native language-based, human-oriented concerns about machine learning framework.\newline

Keywords: \textit{machine learning}; \textit{framework}; \textit{NLP}; \textit{AI ethics}; \textit{explainability}

\section{Introduction}
\label{sec:Introduction}

Two decades ago, some popular algorithms existed and were well documented in scientific literacy, but there was still no easy way to use them. Scientists had to read the equations and the algorithm before implementing it in the desired programming language. Every matrix had to be multiplied, and every derivative had to be computed by the scientist's code. In the last two decades, machine learning has finally flourished. One of the most meaningful frameworks was certainly TensorFlow \cite{tensorflow2015-whitepaper}. This powerful tool helped the community accelerate development and democratize the machine learning field. It helped this field of knowledge reach a more comprehensive range of applicative projects instead of being restricted to academics. 

A few years after the first version of Tensorflow, many others came to the machine learning community. Among the most popular: Scikit-Learn, CNTK, Torch, Matlab, and Keras \cite{wang_various_2019}. In the last few years, a user-friendly framework with a graphical interface named Orange \cite{JMLR:demsar13a} became available, aiming to be even more accessible for the community, especially for the non-expert.  While consistently more accessible over time, requiring less mathematics and fewer programming skills, none of those frameworks has made the ultimate step: the ability to communicate in the native human language.  

Some recent studies compare the most popular machine learning software framework. For instance, framework performances have been recently analysed in \cite{wang_various_2019}. For this same purpose of performance analysis, \cite{verbraeken_survey_2020} divides frameworks into some topics (computational distribution, Tensor Processing Units and Field-Programmable Gate Array (FPGAs)). \cite{zhang_pcamp_2018} compares machine learning frameworks on different hardware platforms, such as Raspberry Pi 3 B+, NVIDIA Jetson, MacBook Pro, Huawei Nexus 6P and Intel FogNode. 

Nguyen et al. in \cite{nguyen_machine_2019} have an essential paper regarding this current research. Their work establishes evaluation criteria for supervised, unsupervised, and reinforcement learning, which are the three prominent families of machine learning. \cite{nguyen_machine_2019} presents an overview of machine learning frameworks and gives the advantages and disadvantages of each. Frameworks are applied in different domains. For instance, \cite{pham_novel_2021} applies it to the Automated Detection of Arrhythmias in ECG Segments, while \cite{motwani_novel_2021} is a framework application in the health domain for smart patient monitoring and recommendation. The work of \cite{noauthor_multidisciplinary_nodate}\cite{agarwal_interpretable_2020} present and compares explainable and interpretable frameworks.

This framework, called AI$^{2}$, proposes a natural language interface. To the authors' best knowledge, there is no machine learning framework offering an Natural language Processing (NLP) interface using a chatbot. This first AI$^{2}$ version proposes an English chatbot, but some other native languages might be proposed later. The NLP domain has flourished recently, especially when using the Transformers technology \cite{rothman2021transformers} \cite{vaswani_attention_2017}. This recent NLP breakthrough created the opportunity to fill the last gap between humans and machine learning frameworks: the ability to communicate in the native human language. This last step has just been done with this proposed AI$^{2}$ framework.  

A state-of-the-art, Transformer-based NLP agent can now correctly interpret users' English requests. Outperforming older methods like \textit{Recurrent Neural Networks} (RNN) \cite{jordan_serial_1986} \cite{noauthor_learning_nodate} and \textit{Artificial Intelligence Markup Language} (AIML) \cite{marietto_artificial_2013}, Transformer technology \cite{karita_comparative_2019} delivers better results. Transformer-based applications exist in multiple domains. For instance, \cite{malhotra_bidirectional_2021} uses it for sentiment analysis.  \cite{guarasci_assessing_2021} evaluates a Transformer's ability to learn Italian syntax. Finally, \cite{chang_design_2022} proposes a chatbot that helps detect and classify fraud in a finance context. 
 
Bidirectional Encoder Representation from Transformers (BERT) \cite{devlin_bert_2019}\cite{rothman_transformers_2021} \cite{adoma_comparative_2020} is a widely used NLP model. It performs exceptionally well when evaluating the context and understanding the intent behind the user's query \cite{ouyang_training_2022}.

Using the BERT NLP model, two pre-trained datasets have been used to build the AI$^{2}$ framework. The first one, BERT (BERT-large), is helpful to answer common questions like "Which dataset has been used?". The second one is RoBERTa (roberta-large) \cite{liu_roberta_2019}. It is only used to answer Yes/No questions like "Is it a clustering problem?". Besides launching the requests, a minor contribution of  AI$^{2}$ is its ability to preprocess the datasets using its NLP chatbot.  

Even if the NLP interface is the main contribution of this paper, other contributions are also proposed.  For instance, another contribution of the AI$^{2}$ framework is the awareness of greenhouse gases (GHG). \textit{CodeCarbon} \cite{lottick_energy_2019} recently proposed a library of functions about GHG awareness and AI$^{2}$ integrates some of those functions and enhances it with machine learning methods. Based on \cite{palacio_xai_2021}, explainability is an essential contribution of this proposed framework. It aims to include ethics principles from the Institute for Ethical AI \& Machine Learning \cite{learning_institute_nodate}. This UK-based research centre develops frameworks that support the responsible development, deployment and operation of machine learning systems. 

\textit{Explainability} is a concept intending to eliminate the "black box" problem. Yoshua Bengio has addressed it, and Judea Perl \cite{goldberg_book_2019}, two Turing awards winners. Over the last decade, ML has reached a certain level of maturity. One of the differences is our expectations of machine learning. There is a need to democratize the methods to non-expert users. Until recently, the scientific community was concerned about lowering the error when using ML algorithms. They were concerned about the performance. Now the expectation is higher. The community still wants good results, but those results have to be found in an explainable, interpretable and ethical context. Human well-being must be the main interest of the ML systems. The results must be explainable. For decades, the "black box" problem was neglected. Now, there are some methods to explain the results and make them understandable to a human. The expectations are also higher regarding the accessibility to the ML methods, GHG awareness and preprocessing. Now, the expectations are higher at different levels. Disposing of the previously presented technologies and based on \cite{AI2_Dessureault_conference}, the contributions of this framework aim to reach expectations with the following targets:  1. democratizing ML frameworks using NLP methods, 2. being GHG aware with a built-in structure to monitor it, 3. being more ethics with a built-in structure systematically explain the results, and 4. having the preprocessing of the data more accessible with an automated NLP based chatbot.  

The following sections of this paper are organized with the following structure: Section \ref{sec:Methodology} describes the proposed methodology. Section \ref{sec:Results} presents the results. Section \ref{sec:Discussions} discusses the results and their meaning, and Section \ref{sec:Conclusion} concludes this research.

\section{Methodology of the AI$^{2}$ framework}
\label{sec:Methodology}

\subsection{Architecture}
\label{sec:sub_Architecture}

\begin{figure}[htb]
\centerline{\includegraphics[width=8cm]{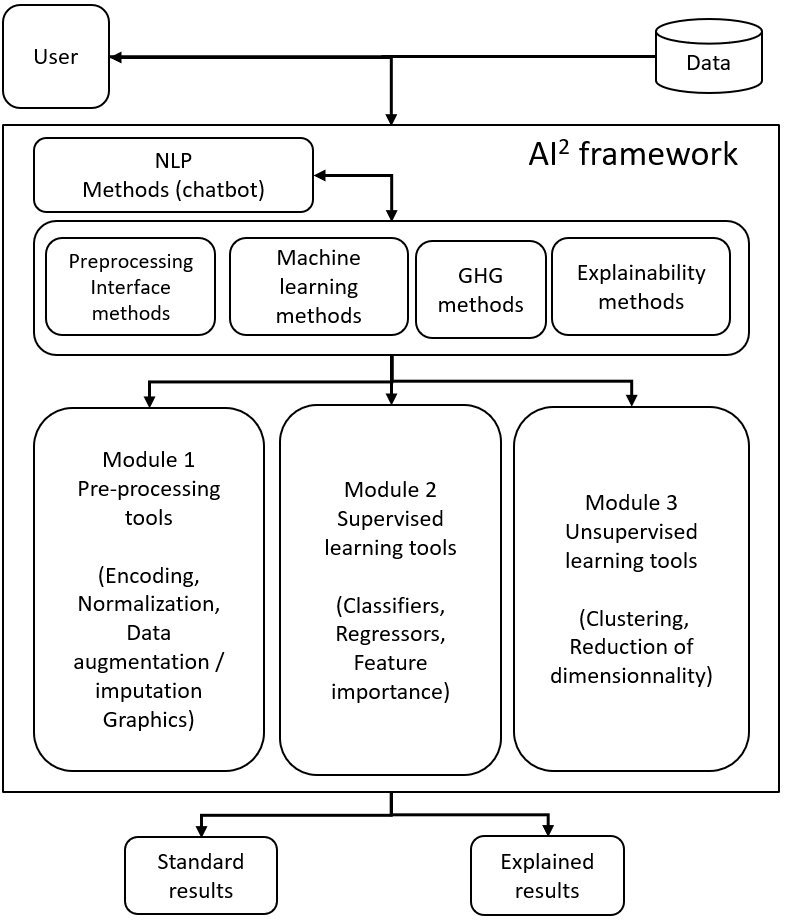}}
\caption{Architecture of the AI$^{2}$ framework.} \label{fig:architecture}
\end{figure}

Fig. \ref{fig:architecture} presents the architecture of the AI$^{2}$ framework. The NLP method, through a chatbot, allows communication with the framework methods and the data using the English language. The kernel of the AI$^{2}$ framework includes four types of methods:  1. Preprocessing methods, 2. Machine learning methods, 3. GHG methods, and 4. Explainability methods.

The preprocessing interface method is done systematically once for each dataset when used for the first time.  The chatbot guides this user throughout the process. It consists of a series of questions to the user about the dataset and each feature/class. The chatbot asks about the type of each field and its normalization method.  
The machine learning methods are the classic supervised and unsupervised learning methods: classifiers, regressors, clustering, dimensionality reduction and a method to evaluate the importance of the features. There are also some new methods like \textit{Decision Process for Dimensionality Reduction} (DPDR) \cite{noauthor_220608974_nodate}, \textit{Decision Process for Dimensionality Reduction before Clustering} DPDRC \cite{Dessureault2_2021}, and CK-Means \cite{noauthor_220608982_nodate}. 
There are also some functions assuring the GHG awareness of the framework. Based on the CodeCarbon library \cite{lottick_energy_2019}, those functions compute the generated GHG for each request. Before launching a request, the GHG functions will predict the GHG generated for this request. They will try to find equivalent requests using clustering methods to save the execution of the subsequent request, thus, saving the generation of GHG. 
The explainability methods offer a complement to the standard machine learning results. The user gets more than the expected results for his request. He gets a well-documented explanation for every result. The form of the explanation varies according to the used algorithm and data. Some examples (like learning curves and the importance of features graphic) are described in the use cases.  The different machine learning methods are divided into three modules. Module 1 includes the preprocessing tools (Encoding, normalization, data augmentation/imputation, graphics). Module 2 consists of the supervised learning tools (classifiers, regressors and the computation of the feature's importance). Finally, module 3 exploits the unsupervised learning tools (clustering and reduction of dimensionality methods). At last, all the results are given in 2 forms: the expected and the explained results. 
AI$^{2}$'s functions can be called without using its NLP interface. Calling the Python function directly without using the English chatbot is very straightforward. The user is responsible for obtaining his own datasets.  No sample dataset is included in this first version of AI$^{2}$. 

\subsection{NLP methods (chatbot)}
\label{sec:sub_NLP_interface}

\begin{figure}[htb]
\centerline{\includegraphics[width=8cm]{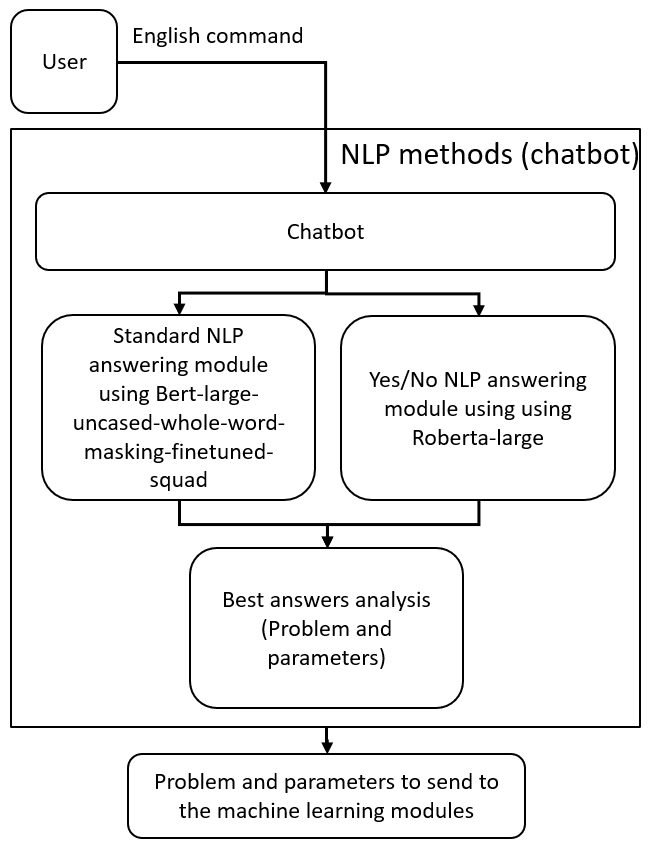}}
\caption{Architecture of the NLP interface in the AI$^{2}$ framework.} \label{fig:nlp_architecture}
\end{figure}

This machine learning framework is its ability to communicate with a user, exploiting a chatbot based on NLP. The chatbot used by the  AI$^{2}$ user interface is made with the Transformers technology, thus, being a state-of-the-art NLP model. In the AI$^{2}$ context, the Transformer technology is used with the "BERT" technology. Fig. \ref{fig:nlp_architecture} presents the NLP architecture of the AI$^{2}$ framework. 

The chatbot uses two types of questions, requiring two different types of NLP pre-trained data. It is essential to note the difference between the datasets that can be processed by the AI's methods and those NLP-based pre-trained datasets used by the chatbot.  \textit{The Standard NLP answering module using Bert-large-uncased-whole-word-masking-finetuned-squad} can help in responding to open questions like: \textit{What is the dataset?}.  As displayed in Table \ref{tab:CSV_File}, this question is associated with the \textit{DATASET} key. The chatbot will try every question having this key to filling out the dataset information. A typical answer to this request can be \textit{iris}, for the \textit{iris} dataset. The pre-trained dataset \textit{Bert-large-uncased-whole-word-masking-finetuned-squad} \cite{devlin_bert_2019} is used to answer this type of question. It is a pretrained model on English language using a Masked Language Modeling (MLM) objective.

Using the \textit{Roberta-large} \cite{liu_roberta_2019} pre-trained dataset, the second type of question is the Yes/No question. A typical question would be \textit{Is this a clustering problem?}. The two possible answers are Yes and No, both associated with a certain level of confidence. As presented in Table \ref{tab:CSV_File}, this question is associated with the \textit{PROBLEM} key and the \textit{CLUSTERING} return value. If the answer to this question is Yes, it will return \textit{CLUSTERING} as an answer to fill out the information. 

As mentioned earlier, every question related to the key is asked in these two types. The NLP system returns an answer for each question and confidence level. The answer related to the best level of confidence is kept. The methodology used to train both pretrained datasets, including the level of confidence formulas are documented in \cite{devlin_bert_2019} and \cite{liu_roberta_2019}. It consists of applying a softmax function on the logits values. The logits variable is known to be the output of a BERT-based Transformer. It is a list of the most probable answers. 

The following describes how the chatbot works. The chatbot first asks \textbf{"Please, enter your English command to the framework"}. The system specifies writing the English command to avoid confounding with a specific programming language-based command used in other frameworks. The expected command is the English instruction to the AI$^{2}$ framework. A typical command could be \textbf{\textit{"I want to perform a clustering using 3 clusters on the iris dataset."}}. From this first answer from  AI$^{2}$, the chatbot will read a Parameters.csv file storing the structure of the required keys, the returned values and the questions to send to the chatbot to access the information. There is no specific order for the keys in this file. The system will request the keys to get the related information. For now, there are 73 rows defined in this file. Those rows designate 19 keys and the questions to access them. Many questions may retrieve each key. It is essential to understand that the framework uses those questions to extract pieces of information from the user command. Those questions are entirely transparent for the users. This file will grow following the new releases of the AI$^{2}$ framework.  Table \ref{tab:CSV_File} presents a sample of this file. \textit{Key} field identifies the information to retrieve. For instance, if AI$^{2}$ seeks the type of problem in the user's command, it will find all the \textit{PROBLEM} rows. It will then interrogate the user's command with all the corresponding \textit{Questions} field. It will keep the answer having to higher level of confidence according to the Transformer. The answer to the question will be returned, except if it is a Yes/No question.  In this case, the \textit{Return value} field will be used.  For instance, if the AI$^{2}$ system replies \textit{Yes} to the question \textit{Is this a clustering problem?} then the returned value will be \textit{CLUSTERING}. The \textit{Type} field indicates \textit{Y/N} for \textit{Yes/No} questions and \textit{Std.} for \textit{standard} questions. 

\begin{minipage}[htb]{1.0\columnwidth}
\captionof{table}{Sample of the Parameters.csv file. Only data used for this example is presented (14 rows on a total of 73).}
\begin{center}
\begin{tabular}{llll}
\hline
\textbf{Key} & \textbf{Type} & \textbf{Return value}& \textbf{Questions to the command} \\
\hline
PROBLEM & Y/N & DIMENSIONALITY & Is this about dimensionality? \\
PROBLEM & Y/N & DIMENSIONALITY & Is this about dimensionality \\ 
 & & & reduction? \\
PROBLEM & Y/N & CLASSIFICATION & Is this about classification? \\
PROBLEM & Y/N & CLASSIFICATION & Is this a classification problem? \\ 
... & ... & ... & ... \\
PROBLEM & Y/N & CLUSTERING & Is this clustering? \\
PROBLEM & Y/N & CLUSTERING & Is this a clustering problem? \\
PROBLEM & Y/N & CLUSTERING & Is this regrouping? \\
PROBLEM & Y/N & CLUSTERING & Is this a regrouping problem? \\
PROBLEM & Y/N & CLUSTERING & Do you want to regroup data? \\
PROBLEM & Y/N & CLUSTERING & Do you want to cluster data? \\
DATASET & Std. &  & What is the dataset? \\
DATASET & Std. &  & Which data are used? \\
NB_CLST & Std. &  & How many groups? \\
NB_CLST & Std. &  & How many clusters? \\
... &  ... & ... & ... \\
\hline
\end{tabular}
\label{tab:CSV_File}
\end{center}
\end{minipage}\newline

Systematically, the chatbot will try to fill the \textit{PROBLEM} key. It must know what kind of problem it is. To find it out, a question list corresponding to the \textit{PROBLEM} key, is processed by the chatbot. If the chatbot can return the answer, the \textit{problem information} (corresponding to \textit{PROBLEM} in the \textit{Key} field, Table \ref{tab:CSV_File}) will be filled. If the first question can return no answer, other questions (corresponding to the key) will be tried to extract information. If no answer can be found after having tried all the questions, the chatbot will prompt to directly ask the user: \textbf{problem to resolve has been found in your text. Please clearly identify the type of problem to solve.}  Then, the algorithm will go to the second and the third required keys: the \textit{DATASET} key and the \textit{NB_CLST} (number of clusters) key. The interface will ask for every crucial information. When the parameter is not mandatory, its default value will be assumed. The same principle is repeated for every required  parameter. An example of a complete sequence is illustrated in Table \ref{tab:Chatbot_sequence}. Remember that the questions are not directly addressed to the user but to his command, aiming to extract meaningful information to execute his request.  

\begin{minipage}[htb]{1.0\columnwidth}
\captionof{table}{Example of a typical command and the question sequence used to extract the information of the command: \textbf{\textit{I want to perform a clustering using 3 clusters on the iris dataset.}}}
\begin{center}
\begin{tabular}{lll}
\hline
\textbf{Questions} & \textbf{Answer}& \textbf{Ret. value} \\
\hline
\multicolumn{3}{c}{\textbf{(To extract the type of the problem)}} \\
Is this about dimensionality? & No & None \\
Is this about dimensionality & No & None \\
reduction? &  &  \\
Is this about classification?  & No & None \\
Is this a classification problem?  & No & None \\
Is this clustering?  & Yes & CLUSTERING \\
\multicolumn{3}{c}{\textbf{(To extract the name of the dataset)}} \\
What is the dataset? & Iris & Iris \\
\multicolumn{3}{c}{\textbf{(To extract the number of clusters)}} \\
How many groups? & (No suitable answer) & None \\
How many clusters? & 3 & 3 \\
\hline
\end{tabular}
\label{tab:Chatbot_sequence}
\end{center}
\end{minipage}\newline

In this example, the answer is \textit{No} for the first four questions, since the command is not about reduction of dimensionality nor classification.  Since the command is about a clustering problem, the answer will be \textit{Yes} to the question \textit{Is this clustering?}. Since the value \textit{Yes} would not mean anything, the corresponding return value (Table \ref{tab:CSV_File}) \textit{CLUSTERING} is returned. After having extracted the problem type, the dataset name is required. The question \textit{What is the dataset?} answer the question. The answer is \textit{Iris}, and the returned value is also \textit{Iris}. The last required information is about the number of clusters needed for the clustering algorithm. There are at least two ways of asking this question since groups and clusters are synonyms. The question \textit{How many groups?} is tried to extract the information. Since the command uses the term \textit{clusters}, no suitable answer is found for this question. The second question will be: \textit{How many clusters?}. The answer and the returned value will be \textit{3}. From this point, AI$^{2}$ has all the required information to launch a clustering algorithm using the Iris dataset and 3 clusters. Some more complete examples are shown in Section \ref{sec:Results}. 

\subsection{Preprocess module}
\label{sec:sub_ModulePreprocess}

The preprocessing method is done systematically once for each dataset when used for the first time. AI$^{2}$ detects when no dataset configuration has been done and stored in a JSON file. The chatbot then asks for the correct configuration for every field, like their name, role in the dataset, and normalization methods. In the end, the dataset's configuration is stored in a JSON file, and the dataset is preprocessed and stored using the same file name, added with a _preprocessed suffix. The chatbot finally asks the user if he wants to process a data imputation of the missing data and a data augmentation. 

\begin{figure}[htb]
\centerline{\includegraphics[width=8cm]{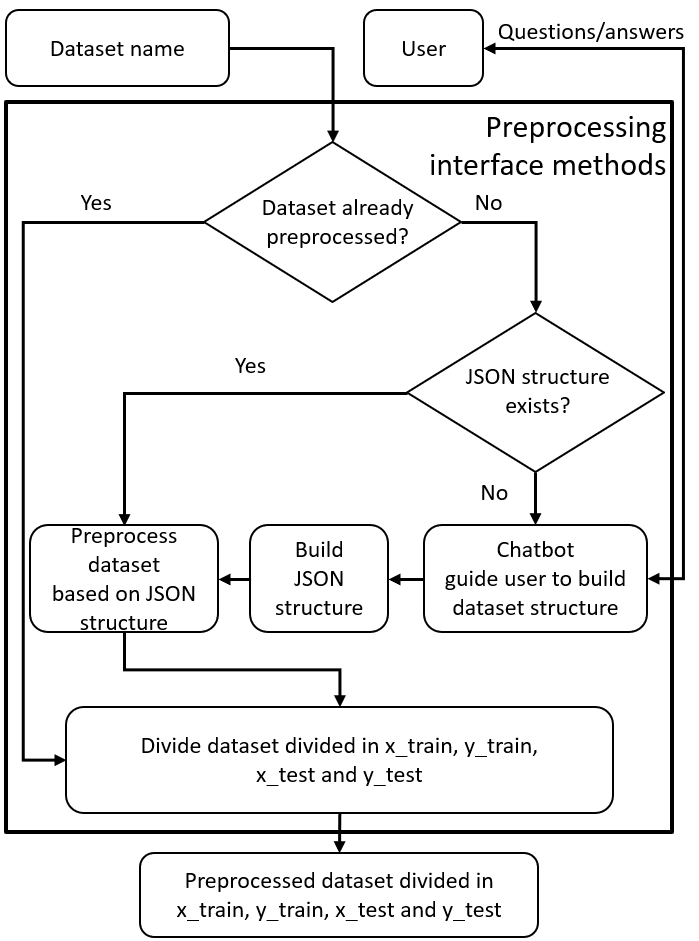}}
\caption{Preprocessing architecture} \label{fig:preprocess_archi}
\end{figure}

Fig. \ref{fig:preprocess_archi} presents the functionalities of the preprocess modules. First, a dataset name is given to the module. If a preprocessed version of the dataset already exists, the module will open it, dividing it into train and test data. If the preprocessed files do not exist, the system will try to find the corresponding JSON file. If the JSON file exists, the system will use it to build the preprocessed file. If it does not exist, the AI$^{2}$ chatbot will guide the user through some questions about the field and create the final JSON file containing the structure of the dataset, and it will create the preprocessed dataset from this JSON file. Ultimately, it will also split the data into train and test data. 

Fig. \ref{fig:JSON} shows an example of a structure configuration JSON file. The included fields in JSON format are the following: \textit{dataset_name} is the name of the dataset. \textit{dataset_description} is a description of the dataset. \textit{feat_no} is the number of the feature. \textit{feat_label} is the label given to this feature. The type of the feature is given by \textit{feat_type}. Possible values are 1. Feature field, 2. Regression value field, 3. Class field, and 4. Class for neural network field (to be one-hot encoded). The last field is \textit{feat_normalization}. Possible values are 1. No normalization, and 2. MinMax normalization. 

\begin{figure}
\begin{Verbatim}[frame=single]
iris.json
{
    "dataset_name": "iris",
    "dataset_description": "iris dataset",
    "feat_no": [
        0,
		... 
    ],
    "feat_label": [
        "Petal length in cm",
		...
    ],
    "feat_type": [
        "1",
		...
    ],
    "feat_normalization": [
        "1",
		...
    ]
}
\end{Verbatim}
\caption{iris.json structure file}
\label{fig:JSON}
\end{figure}

Fig. \ref{fig:Preprocess} shows an example of an exchange between the chatbot and the user, aiming to proprocessing the data. 

\begin{figure}[htb]
\begin{Verbatim}[frame=single]
Let us preprocess the iris dataset. Please, answer the 
  following questions:
What is the description of the iris dataset (ENTER to skip)?
>This dataset describes the features and the class of the iris 
  dataset.
  
What is the name of the field 0? (Value example: 5.1)
>Sepal length in cm

What is the type of field Sepal length in cm? (1. Feature 2. 
  Predicted value 3. Class  4. Class (to be converted ONE-HOT 
  for neural network)
>1

What is the normalization applied to Sepal length in cm? 
 (1. None 2. MinMax)
>1

       (... And so on for each feature and class.)

Saving dataset configuration...
The configuration is saved to iris.json
Processing to the file conversion...
The configuration is saved to iris_preprocessed.csv
\end{Verbatim}
\caption{An example of the exchange between the chatbot and the user for the data preprocessing.}
\label{fig:Preprocess}
\end{figure}

\subsection{Machine learning methods}
\label{sec:sub_ML_Methods}

Any framework requires a tremendous amount of development hours. 
This framework is still in development, yet it has some contributions to bring to the scientific community. Some known algorithms are included, resolving most machine learning problems (prediction, classification, and others). Table \ref{tab:ML_methods} shows algorithms included in AI$^{2}$.   

\begin{minipage}[htb]{1.0\columnwidth}
\captionof{table}{Machine learning algorithms are included in the AI$^{2}$ framework.}
\begin{center}
\begin{tabular}{lll}
\hline
\textbf{No.} & \textbf{Modules}& \textbf{Algorithms} \\
\hline
1. & Pre-processing 		& IQR  \\
2. &						& SMOTE \\
3. & 						& KNNImputer \\
4. & 						& xGEWFI metric  \\
5. & Supervised learning	& Neural network regressor \\
6. &    					& Neural network classifier  \\
7. & 						& Random Forest \\
8. & Unsupervised learning	& K-Means  \\
9. & 						& CK-Means  \\
10. & 						& Silouette metric \\
11. & 						& PCA \\
12. & 						& DPDRC \\
13. & 						& DPDR \\
14. & 						& FRSD \\
\hline
\end{tabular}
\label{tab:ML_methods}
\end{center}
\end{minipage}\newline

The pre-processing methods (Module 1) are regrouped into one callable function. This function can do the whole process of finding the outliers, augmenting the data and imputing the missing data. The recent explainable metric named xGEWFI \cite{noauthor_220608980_nodate} is used to evaluate the performance of the data generation (imputation and augmentation). It considers the importance of the feature and each feature error to evaluate the global error of the data generation process. Inter Quartile Range (IQR) algorithm is used to find the outliers. Data generation (augmentation and imputation of missing data) are made with a SMOTE algorithm \cite{chawla_smote_2002} and a KNNImputer \cite{troyanskaya_missing_2003}, respectively. 

Some neural networks (multilayer perceptron doing regressions and classifications) \cite{ramchoun_multilayer_2016} are available for supervised learning functions (Module 2). A Random Forest (RF) algorithm \cite{biau_random_2016} is used as a classifier and regressor. It is also used to evaluate the importance of the features. 

Some unsupervised learning methods (Module 3) are also available. The K-means algorithm \cite{ahmed_k-means_2020} can be executed for clustering problems. The CK-Means algorithm \cite{noauthor_220608982_nodate} can be called to extract data from the cluster's intersection. The metric to evaluate the cluster consistencies of those first two algorithms is the Silhouette Index (SI) \cite{rousseeuw_silhouettes_1987}. Concerning the dimensionality reduction, the Principal Component Analysis (PCA) algorithm \cite{jolliffe_principal_2016} is included in the AI$^{2}$ framework. 
Two new decision processes are also included to help with the dimensionality reduction problems. 1. Decision Process for Dimensionality Reduction before Clustering (DPDRC) \cite{Dessureault2_2021} and 2. Decision Process for Dimensionality Reduction (DPDR)  \cite{noauthor_220608974_nodate}. Those two are used in unsupervised learning and supervised learning contexts, respectively. In an unsupervised learning context, Feature Ranking Process Based on Silhouette Decomposition (FRSD) \cite{yu_ensemble_2020} helps evaluate the importance of the features. 

\subsection{GHG Methods - CodeCarbon integration in AI$^{2}$}
\label{sec:sub_CodeCarbon}

\begin{figure}[htb]
\centerline{\includegraphics[width=8cm]{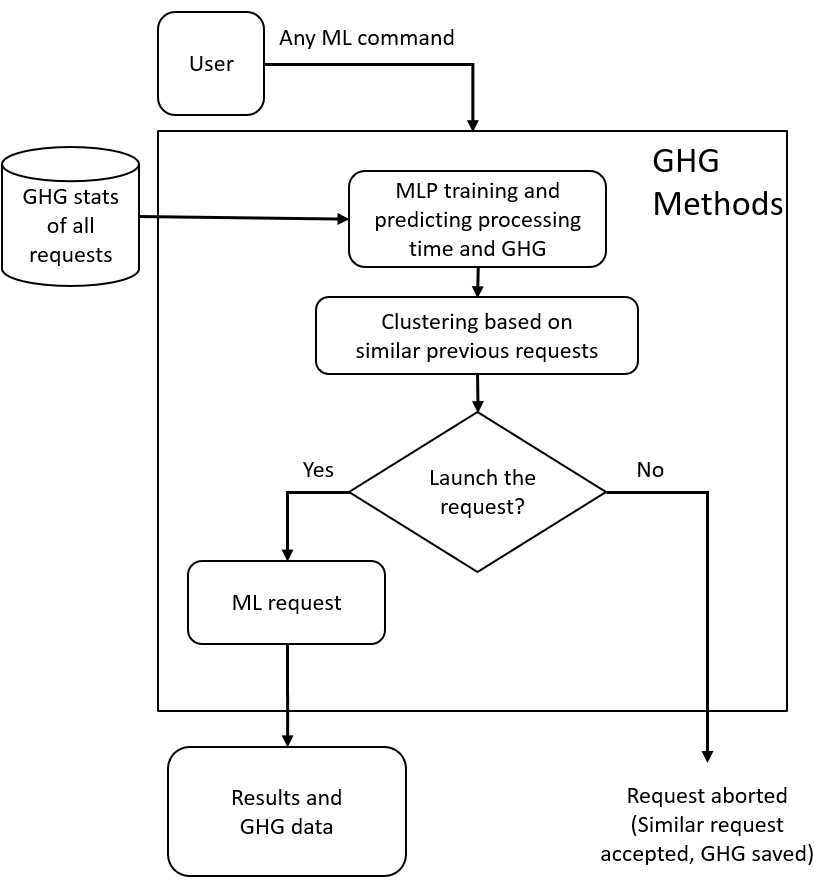}}
\caption{GHG module architecture} \label{fig:GHG_archi}
\end{figure}

Climate change is an essential issue for humanity. It is our responsibility to be aware of it and to do everything that can be done to contribute to lower GHG. We know that computer sciences, particularly machine learning, can significantly generate GHG while executing on CPU and GPU. The CodeCarbon library is an important initiative available to data scientists, so they can be aware of their impact on GHG. The following quote can be found on the CodeCarbon website (at pypi.org/project/codecarbon/) based on \cite{lottick_energy_2019}: \textit{While computing currently represents roughly 0.5\% of the world’s energy consumption, that percentage is projected to grow beyond 2\% in the coming years, which will entail a significant rise in global CO2 emissions if not done properly. Given this increase, it is important to quantify and track the extent and origin of this energy usage, and to minimize the emissions incurred as much as possible. For this purpose, we created CodeCarbon, a Python package for tracking the carbon emissions produced by various kinds of computer programs, from straightforward algorithms to deep neural networks. By taking into account your computing infrastructure, location, usage and running time, CodeCarbon can provide an estimate of how much CO2 you produced, and give you some comparisons with common modes of transportation to give you an order of magnitude.
} The contribution of this paper is to embed this library's features in a machine learning framework, add some machine learning-based functions to predict the subsequent request amount of GHG, and try to spare its execution by proposing some alternatives. Fig. \ref{fig:GHG_archi} explains those embedded  GHG functionalities.

First, every GHG statistic (request name, machine learning algorithm used, dataset, number of data, fields, elapsed time, GHG emissions) is stored in a file. When a user is about to launch a new request, from this stored historic, AI$^{2}$ framework will try to predict the amount of GHG this subsequent request will generate. A multilayer perceptron (MLP) is used to evaluate this GHG amount. This MLP have 5 hidden layers of 25 neurons. It uses a \textit{relu} activation function and an \textit{adam} solver.  Then, a k-means clustering algorithm is used to regroup every similar request to the current request. The list is proposed to the user so he can spare his execution, with some similar results available from the historic. Knowing how much GHG will be generated and knowing the similar results of the past, the user will finally decide if yes or no he wants to execute his new request. Fig. \ref{fig:clust_similar_request} presents an example of the information from the chatbot concerning the GHG before launching a new request.

\begin{figure}[htb]
\begin{Verbatim}[frame=single]
Predicted execution time (in sec): 4.498
Predicted generated GHG: 4.899e-05 kg CO2

Here are the most similar requests in case launching another 
request can be avoided. 
   Request _2022-11-21_21-23-43 using dataset make_blob
   Request _2022-11-22_13-54-45 using dataset make_blob
   Request _2022-11-22_14-29-32 using dataset make_blob
Launch the request (y/n)? 
\end{Verbatim}
\caption{Information from the chatbot concerning the GHG before launching a new request.}
\label{fig:clust_similar_request}
\end{figure}

\subsection{Explainability methods}
\label{sec:sub_Explainability}

The goal of this part is to get rid of the famous "black box" problem in machine learning. When most frameworks usually display the results for every executed algorithm, AI$^{2}$ will systematically display the ad-hoc graphics, tables and texts that will ensure a better explainability for a particular algorithm. It could be some learning curves, some scalability curves, and some confusion matrices. For instance, for a clustering process, some stacked radar graphics (one per cluster) are produced, plus a Silhouette index graphic that shows the cluster's consistency. A cluster table and a text (in LaTeX format) are also created to complete the explainability of the process. For each machine learning algorithm, the totality of the graphics, tables and texts are generated using the \textit{explain()} method.  

\section{Results}
\label{sec:Results}

The following presents five functional use cases. They emphasise the singularity of the AI$^{2}$ framework. It shows how a user can execute some requests to this framework and what type of results are presented as output. The output graphics, tables, and texts are not presented in this paper for two reasons: 1. It is not what this paper intends to demonstrate. For instance, there is no need to show result for a simple clustering K-mean process. 2. There would have needed too many graphics, tables and texts to present in this paper. Case 1 to case 5 present a clustering, a reduction of dimensionality, a classification, a prediction, and an evaluation of the feature's importance.   
    
\subsection{Case 1: Clustering}
\label{sec:sub_results_case1}
                     
The first case is about a clustering process. As mentioned earlier, the user must write his query in English in the chatbot. For this first case, the following command has been entered: \textit{I want to perform a clustering using iris dataset and having 3 clusters.}

From the Parameters.csv file where a sample is presented in Table \ref{tab:Table_case1}, the following questions (Table \ref{tab:Table_case1}) will be generated by the chatbot to fill the required information about a clustering process : \newline

\begin{minipage}[htb]{1.0\columnwidth}
\captionof{table}{Required information and questions to access it.}
\begin{center}
\begin{tabular}{llll}
\hline
\textbf{Key} & \textbf{Type} & \textbf{Return value}& \textbf{Questions} \\
\hline
PROBLEM & Y/N & CLUSTERING & Is this clustering? \\
PROBLEM & Y/N & CLUSTERING & Is this a clustering problem? \\
PROBLEM & Y/N & CLUSTERING & Is this regrouping? \\
PROBLEM & Y/N & CLUSTERING & Is this a regrouping problem? \\
PROBLEM & Y/N & CLUSTERING & Do you want to regroup data? \\
PROBLEM & Y/N & CLUSTERING & Do you want to cluster data? \\
DATASET & Std. &  & What is the dataset? \\
DATASET & Std. &  & Which data are used? \\
NB_CLST & Std. &  & How many clusters? \\
NB_CLST & Std. &  & How many groups? \\
\hline
\end{tabular}
\label{tab:Table_case1}
\end{center}
\end{minipage}\newline

At this first step, AI$^{2}$ transparently tries to find the answers in the command entered by the user. After this first step, if AI$^{2}$ misses some information, the chatbot will ask for it until every critical information is defined. From this example, the iris dataset is loaded, a \textit{k-means} algorithm is launched with the parameter $n_clusters=3$ and using the default parameters $random_state=1$ and $init="k-means++"$.  

The primary results are displayed, presenting a data table along with their clusters, that what most of the frameworks would do. Using AI$^{2}$, each graphic, table and text can be called using the \textit{explain()} method. In this first case, stacked radar graphics are generated for each cluster, allowing to visualize the profile of every cluster. It also generates a graphic of the Silouhette Index, showing and measuring the consistency of every cluster, and finding the mean of the whole clustering process. For each table and graphic, a short text describing it is generated in LaTeX format.

\subsection{Case 2: Reduction of dimensionnality}
\label{sec:sub_results_case2}
The second case is about the reduction of dimensionality. The entered command was: \textit{reduction of dimensionality with iris dataset and having 3 components.}  The only required parameter is the targeted number of components that should be used to downsize the dataset. If this parameter is not specified in the command, the chatbot will directly ask to specify it. Since it is defined in this case command, AI$^{2}$ will extract three components of the dataset using the PCA algorithm. Always from the Parameters.csv file, the questions shown in Table \ref{tab:Table_case2} will be generated by the chatbot to fill the required information about a reduction of dimensionality process : \newline

\begin{minipage}[htb]{1.0\columnwidth}
\captionof{table}{Required information and questions to access it.}
\begin{center}
\begin{tabular}{llll}
\hline
\textbf{Key} & \textbf{Type} & \textbf{Return value}& \textbf{Questions} \\
\hline
PROBLEM & Y/N & DIMENSIONALITY & Is this about dimensionality? \\
PROBLEM & Y/N & DIMENSIONALITY & Is this about dimensionality \\
 &  &  & reduction? \\
PROBLEM & Y/N & DIMENSIONALITY & Is this about reduction  \\
 &  &  & of dimensionality? \\
PROBLEM & Y/N & DIMENSIONALITY & Is this a regrouping problem? \\
PROBLEM & Y/N & DIMENSIONALITY & Is this a dimensionality problem? \\
PROBLEM & Y/N & DIMENSIONALITY & Is this a dimensionality  \\
 &  &  & reduction problem? \\
DATASET & Std. &  & What is the dataset? \\
DATASET & Std. &  & Which data are used? \\
NB_CMPS & Std. &  & How many components? \\
\hline
\end{tabular}
\label{tab:Table_case2}
\end{center}
\end{minipage}\newline

The result is a dataset having three principal components (reduced with the PCA algorithm). The \textit{explain()} method generated two graphics: 1. the covariance heatmap of the initial features. 2. a bar graph of the three extracted features' importance (explained variance ratio). For both graphics, a short LaTeX explaining it is generated. 

\subsection{Case 3: Classification}
\label{sec:sub_results_case3}
The following case is about the typical problem of classification. For this case, a multiple sentences English is given: \textit{Perform a classification of the iris dataset. I want this request to be reproducible. Test [4.8,3.0,1.4,0.2] value.}  The first sentence of the command is straightforward. Those two sentences are written in a single command. It calls a classification of the iris dataset. To do so, it will call a multilayer perceptron (MLPClassifier from the Scikit-learn framework). The second sentence mention that it requires reproducible results. This will set the seed of the \textit{random_state} parameter to the "1" integer value, assuring the request gives the same result every time. The opposite would have been a "random request". The seed would have been set to \textit{None}, allowing the request to give slightly different results due to some random synaptic connection initialization. If it is not specified, the request is reproducible. The final sentence commands to try some values. In other words, it aims to classify the specified values [4.8,3.0,1.4,0.2]. The questions in Table \ref{tab:Table_case3} will be extracted from the text command.\newline

\begin{minipage}[htb]{1.0\columnwidth}
\captionof{table}{Required information and questions to access it.}
\begin{center}
\begin{tabular}{llll}
\hline
\textbf{Key} & \textbf{Type} & \textbf{Return value}& \textbf{Questions} \\
\hline
PROBLEM & Y/N & CLASSIFICATION & Is this about classification? \\
PROBLEM & Y/N & CLASSIFICATION & Is this a classification problem? \\
PROBLEM & Y/N & CLASSIFICATION & Do you want to classify data? \\
DATASET & Std. &  & What is the dataset? \\
DATASET & Std. &  & Which data are used? \\
RANDOM & Y/N & RANDOM & Is this a random request? \\
RANDOM & Y/N & REPRODUCTIBLE & Is this a reproductible request? \\
TEST & Std. & & What are the test values? \\
TEST & Std. & & What values do you want \\
 &  & & to be tested? \\
\hline
\end{tabular}
\label{tab:Table_case3}
\end{center}
\end{minipage}\newline

The classification result will then be shown. The training is done with cross-validation having the parameter k = 10. The whole dataset is split k times, and the subsets are used to validate to process. The training and validation scores are returned for each step of the cross-validation. While both scores are increasing, the training may continue the learning process. When the training score is still increasing while the validation score starts to decrease, it is precisely the right time to stop the training process. Stopping before that moment creates under-fitted training, and stopping after that point results in overfitted training. Calling the \textit{explain()} method, a learning curve is generated of both the training score and validation score based on the cross-validation. 

A state-of-the-art method executes the neural network to classify the data. Earlier in the process, the train and the test data were split, allowing the algorithm to train and evaluate the performances. Performance graphics is also created, showing the performance of the training. Scalability graphics show the ratio of the number of processed data/processing time. Like the other cases, LaTeX texts are generated to explain every graphic. 

\subsection{Case 4: Prediction}
\label{sec:sub_results_case4}

This case aims to demonstrate the prediction feature of the AI$^{2}$ framework, using the \textit{MLPRegressor} from \textit{Scikit-learn}. It also shows how to preprocess a dataset before calling an algorithm. This preprocessing can be called in the chatbot. In this case, the following English command is given: \textit{Do the preprocess of the iris2 dataset.} Note that the iris2 dataset is identical to the iris dataset, except that the class field is not included. Selecting the columns of a dataset is not included in this first version of AI$^{2}$, but it will be in a different version. The iris2 dataset remains with four features: Sepal length, sepal width, petal length and petal width. The value of the petal width must be predicted. When responding to the chatbot's questions, the user must specify that the first three fields are non-normalized features and the fourth is a regression value. After responding to the questions in the chatbot, the iris2.json file is created, containing the information about the configuration. The iris2_preprocessed.csv data file is also created containing the preprocessed data. A second command can be sent to AI$^{2}$ using the chatbot: \textit{I want to make a prediction using the iris dataset. Test [4.5,3.1,1.2]}. The questions in Table \ref{tab:Table_case4} will be extracted from the text command.\newline

\begin{minipage}[htb]{1.0\columnwidth}
\captionof{table}{Required information and questions to access it.}
\begin{center}
\begin{tabular}{llll}
\hline
\textbf{Key} & \textbf{Type} & \textbf{Return value}& \textbf{Questions} \\
\hline
PROBLEM & Y/N & PREDICTION & Do you want to make \\
& &  &  a prediction? \\
PROBLEM & Y/N & PREDICTION & Is this a prediction problem? \\
PROBLEM & Y/N & PREDICTION & Do you want to predict something? \\
DATASET & Std. &  & What is the dataset? \\
DATASET & Std. &  & Which data are used? \\
TEST & Std. & & What are the test values? \\
TEST & Std. & & What values do you want \\
 &  & & to be tested? \\
\hline
\end{tabular}
\label{tab:Table_case4}
\end{center}
\end{minipage}\newline

Three graphics are generated to explain the results as in \ref{sec:sub_results_case3}. A learning curve is displayed to ensure no training underfitting or overfitting. A second graphic shows the performance of the training process. Moreover, a third graphic shows the scalability of the training. As always, LaTeX texts are created to explain the figures, ready to be cut and pasted in a LaTeX document.  

\subsection{Case 5: Feature's importance}
\label{sec:sub_results_case5}

This next case shows how to evaluate the feature importance in the AI$^{2}$ framework. The following command has been typed in AI$^{2}$'s chatbot: \textit{Find the importance of the features with the iris dataset.} This command calls a Random Forest algorithm. More precisely, the \textit{RandomForestClassifier} and the \textit{RandomForestRegressor} from the \textit{Scikit-learn} framework. According to the configuration file's content (iris.json in this case), it will detect whether it is a dataset made for regression or classification. In this case, the iris is a dataset made for classification, so the \textit{RandomForestClassifier} algorithm will be used. From the Parameters.csv file, the questions shown in Table \ref{tab:Table_case5} are asked by the chatbot to fill in the information about the feature importance algorithm: \newline

\begin{minipage}[htb]{1.0\columnwidth}
\captionof{table}{Required information and questions to access it.}
\begin{center}
\begin{tabular}{llll}
\hline
\textbf{Key} & \textbf{Type} & \textbf{Return value}& \textbf{Questions} \\
\hline
PROBLEM & Y/N & FEAT_IMP & Is this about feature importance? \\
PROBLEM & Y/N & FEAT_IMP & Is this about the importance  \\
& & & of the features? \\
PROBLEM & Y/N & FEAT_IMP & Is this a feature importance problem? \\
PROBLEM & Y/N & FEAT_IMP & Do you want to know the  \\
 &  &  & feature importance? \\
DATASET & Std. &  & What is the dataset? \\
DATASET & Std. &  & Which data are used? \\
\hline
\end{tabular}
\label{tab:Table_case5}
\end{center}
\end{minipage}\newline

The \textit{explain()} method gives a graphic where the X axe represents the index of the features, and the Y axe shows each feature's normalized level of importance.  A LaTeX explanation text is generated as usual. 

\subsection{GHG algorithms validation}
\label{sec:sub_results_validation}

\begin{figure}[htbp]
\centerline{\includegraphics[width=8cm]{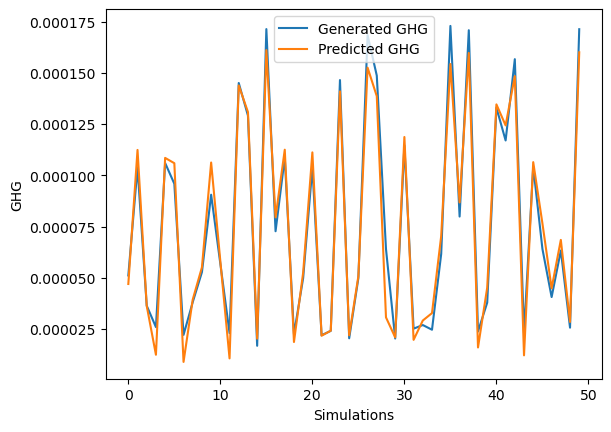}}
\caption{Validation of the predicted and real GHG} \label{fig:valid_GHG}
\end{figure}

As stated in \ref{sec:sub_CodeCarbon}, the AI$^{2}$ framework predicts GHG for each algorithm to be executed. Execution time is also predicted before calling the machine learning algorithm.
To validate those predictions, a clustering algorithm has been called within 50 iterations loops. For each execution, a random-sized dataset of 10,000 to 50,000 rows and 5 to 20 features have been used. Those datasets were generated by the \textit{make_blob()} function of the \textit{scikit-learn} framework. Fig. \ref{fig:valid_GHG} shows the validation of the predicted and real values of the generated GHG. X axe displays the 50 iterations, and the Y axe shows the level of GHG (in $kg CO_2$ unit). The regression algorithm was trained from a dataset containing 1382 rows containing the request's historical.   

\begin{figure}[htb]
\centerline{\includegraphics[width=8cm]{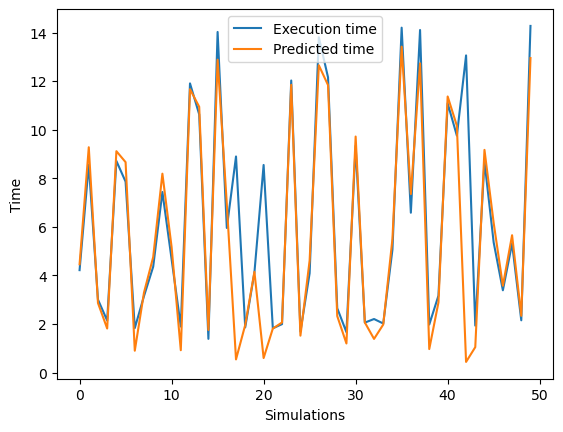}}
\caption{Validation of the predicted and real execution time} \label{fig:valid_time}
\end{figure}

Fig. \ref{fig:valid_time} displays the validation of the predicted and actual values of the execution time for every iteration of the loop. X axe shows the 50 iterations, and the Y axe shows the execution time (in sec.).

Concerning the predicted and real GHG and execution time, it can be seen that the signal is reasonably reconstructed. 

Finally, before launching each request, AI$^{2}$ proposes similar requests from the request's historic after extracting this information using a clustering process. Fig. \ref{fig:clust_similar_request} presents an example of the AI$^{2}$ propositions of the similar requests. 

\begin{figure}
\begin{Verbatim}[frame=single]
Here are the most similar requests in case launching another request can be avoided. 
   Request _2022-11-21_21-23-43 using dataset make_blob
   Request _2022-11-22_13-54-45 using dataset make_blob
   Request _2022-11-22_14-29-32 using dataset make_blob
\end{Verbatim}
\caption{iris.json structure file}
\label{fig:clust_similar_request}
\end{figure}

\section{Discussions}%
\label{sec:Discussions}%

The first contribution of this paper is to present an accessible framework. With its state-of-the-art NLP methods, this machine learning framework is a pioneer in communicating with a non-expert user in English. The new Transformers technology allows the AI$^{2}$ framework to receive native language commands extracted, parsed and executed. When there is an essential missing parameter, AI$^{2}$ will use its chatbot to communicate with the user, asking him to enter the missing information. With this NLP interface, a user can exploit the AI$^{2}$ framework without knowing how to code with a programming language like Python or others. 
 
The AI$^{2}$ framework is GHG-aware, and this is the second contribution of this paper. The CodeCarbon library is encapsulated in each of its ML functions, allowing the calculation of the GHG for each algorithm executed. Those GHG records are kept in a register and used to predict, based on ML, the GHG generated before the execution. AI$^{2}$ also propose some similar registered requests, also based on ML, to save this execution and save GHG. 

The opposite of most other frameworks, AI$^{2}$ systematically encapsulates the most important format of explanations about the data and the results. This aspect of the framework is crucial to solving the famous black-box problem. This is the third contribution of this paper. Most of the machine learning framework is not systematically offering some explainability with the results. AI$^{2}$ does. It generates, for each request, some graphics, some tables, and some texts explaining the results and the data, thus, making this framework more ethical than others.  
 
The final contribution of this paper is data preprocessing. It usually takes time to code a suitable preprocessing of the data. The AI$^{2}$ framework proposes a method based on communication with the chatbot to automatize this process. Guided by the AI$^{2}$ chatbot, the user may do some basic preprocessing of its datasets by establishing the dataset's structures. Having a structure stored in a JSON file, the preprocessing module can generate a new preprocessed dataset. 

Comparing AI$^{2}$ with other machine learning frameworks, what is the advantage of using it? For now, there are frameworks more complete and more sophisticated. The AI$^{2}$ framework targets non-expert users who need a machine-learning algorithm to process their data. Typical AI$^{2}$ users would be, for instance, researchers, engineers, teachers and students in natural science, and so on. A significant part of the scientific community cannot program complex algorithms using a programming language. An NLP interface is the best solution since it requires no programming skills. 

Table \ref{tab:comparison} shows a comparison between AI$^{2}$ and the other popular machine learning framework, according to 3 criteria: 1. NLP interface, 2. GHG awareness, 3. Explainability, and 4. NLP Preprocessing.  

\begin{minipage}[htb]{1.0\columnwidth}
\captionof{table}{Comparison of the popular machine learning frameworks, specialized frameworks, and AI$^{2}$}
\begin{center}
\begin{tabular}{lllllll}
\hline
\textbf{Framework} & \textbf{NLP} & \textbf{GES} & \textbf{Explain.} & \textbf{Prepro.}  & \textbf{Code}  & \textbf{Ref.}  \\
 &  & \textbf{Aware} &  &   & \textbf{req.} &  \\
\hline
AIX360  & NO & NO & \textbf{YES} & NO & YES & \cite{agarwal_interpretable_2020} \\
ELI5  & NO & NO & \textbf{YES} & NO & YES & \cite{agarwal_interpretable_2020}\\
Gluon  & NO & NO & NO & NO & YES & \cite{nguyen_machine_2019}\\
Keras  & NO & NO & NO & NO & YES & \cite{nguyen_machine_2019}\\
LIME   & NO & NO & \textbf{YES} & NO & YES & \cite{noauthor_multidisciplinary_nodate}\\
Matlab  & NO & NO & NO & NO & YES & \cite{nguyen_machine_2019}\\
MXNet  & NO & NO & NO & NO & YES & \cite{zhang_comparison_2017}\\
Orange  & NO & NO & NO & NO & \textbf{NO} & \cite{JMLR:demsar13a}\\
PyTorch  & NO & NO & NO & NO & YES & \cite{nguyen_machine_2019} \\
Scikit-learn  & NO & NO & NO & NO & YES & \cite{nguyen_machine_2019}\\
SHAP  & NO & NO & \textbf{YES} & NO & YES & \cite{noauthor_multidisciplinary_nodate}\\
Skater   & NO & NO & \textbf{YES} & NO & YES & \cite{agarwal_interpretable_2020}\\
Tensorflow  & NO & NO & NO & NO & YES & \cite{zhang_comparison_2017}\\
What-if Tool  & NO & NO & \textbf{YES} & NO & YES & \cite{noauthor_multidisciplinary_nodate} \\
XAI & NO & NO & \textbf{YES} & NO & YES & \cite{noauthor_multidisciplinary_nodate}\\
CodeCarbon & NO & \textbf{YES} & NO & NO & NO & \cite{lottick_energy_2019}\\
\textbf{AI$^{2}$}  &\textbf{YES} &\textbf{YES} &\textbf{YES} &\textbf{YES} &\textbf{NO} & \cite{AI2_Dessureault_conference} \\
\hline
\end{tabular}
\label{tab:comparison}
\end{center}
\end{minipage}\newline

Note that some well-known frameworks may seem absent from the list: CNTK and Theano are no longer supported. Caffe2 is merged with PyTorch. According to Table \ref{tab:comparison}, we can regroup the frameworks into three categories: 1. The general, multi-purpose frameworks (Gluons, Keras, MXNet, Tensorflow, PyTorch, Matlab, Orange and Scikit-learn) 2. The Explainability frameworks (AIX360, ELI5, LIME, SHAP, Skater and XAI), and 3. The GHG-aware framework (CodeCarbon). 

This table shows AI$^{2}$'s novelty. It is the only framework that combines all the studied criteria (NLP interface, GES awareness, Explainability, Preprocessing, and Coding required). It is the first framework to have an NLP interface to send the instructions to the framework. Several frameworks integrate the explainability of the data and the models, but no general and multi-purpose framework includes it. AI$^{2}$: The next leap toward native language-based, GHG-aware and explainable ML framework.

\section{Conclusion}%
\label{sec:Conclusion}%
This framework proposes a tool for the non-expert to use machine learning methods. It offers an NLP interface so the user can communicate with the framework using a chatbot. It encapsulates some very concrete functions to provide ecological awareness. It includes the principle of explainability, proposing expanded results explications for different algorithms. It finally allows preprocessing of data using an English chatbot. 

This framework could be the first draft of a long series of improvements. There are many future works to do for each of its contributions. Regarding its NLP interface, this framework can be improved by training the pre-trained Transformer on a specific machine learning-oriented text corpus. Likely, the NLP's performance will significantly improve. The chatbot method can also be optimized to minimize errors and recognize the user's intentions. Questions used to extract command information can be improved by increasing the quality and the number of questions. GHG awareness can be improved. Better methods can be found to minimize wasted energy, maximize the GHG estimation before calling an algorithm, and cluster similar requests. There is a lot to do, but this framework has the merit of being aware of the climate change problem and proposing a modest solution. Explanations available for each data and machine learning algorithm can also be optimized in quantity and quality. Some essential explanations are included in this framework, but those need to be systematically included. Regarding the preprocessing module, there are many things to add. For instance, some normalization methods can be added. The rows and columns selection can be added to this module, also. Some graphics can be added to plot data at the preprocessing stage. Finally, this framework contains a limited number of ML algorithms. Some more ML algorithms can be easily added to the AI$^{2}$ framework.

\section{Acknowledment}%
\label{sec:Acknowledment}%
This work has been supported by the "Cellule d’expertise en robotique et intelligence artificielle" of the Cégep de Trois{-}Rivières and the Natural Sciences and Engineering Research Council.

\bibliographystyle{plain}%
\bibliography{paper}

\end{document}